%% file: main.tex
\documentclass{ecai} 

\usepackage{pdfpages}
\usepackage{appendix}
\usepackage{latexsym}
\usepackage{amssymb}
\usepackage{amsmath}
\usepackage{amsthm}
\usepackage{amsfonts}
\usepackage{booktabs}
\usepackage{enumitem}
\usepackage{multirow}
\usepackage{graphicx}
\usepackage{color}
\usepackage{todonotes}
\usepackage{subfigure}
\usepackage{bm}

\usepackage[switch]{lineno}
\usepackage{siunitx,tabularx,ragged2e}
\newcolumntype{Y}{>{\RaggedRight\arraybackslash}X}
\newcolumntype{T}[1]{S[table-format=#1]}

\usepackage{algorithm}
\usepackage{algorithmicx,algpseudocode}
\def\NoNumber#1{{\def\alglinenumber##1{}\State #1}\addtocounter{ALG@line}{-1}}
\usepackage[switch]{lineno}

\usepackage{listings}
\definecolor{codegreen}{rgb}{0,0.6,0}
\definecolor{codegray}{rgb}{0.5,0.5,0.5}
\definecolor{codepurple}{rgb}{0.58,0,0.82}
\definecolor{backcolour}{rgb}{0.95,0.95,0.92}
\definecolor{backcolour}{rgb}{0.95,0.95,0.92}
\definecolor{codegreen}{rgb}{0,0.6,0}
\definecolor{backcolour}{rgb}{0.95,0.95,0.92}
\definecolor{codegreen}{rgb}{0,0.6,0}
\definecolor{codegreen}{rgb}{0,0.6,0}
\definecolor{codegray}{rgb}{0.5,0.5,0.5}
\definecolor{codepurple}{rgb}{0.58,0,0.82}
\definecolor{backcolour}{rgb}{0.95,0.95,0.92}
\definecolor{gray1}{HTML}{E6E6E6}

\definecolor{backcolour}{rgb}{0.95,0.95,0.92}
\definecolor{codegreen}{rgb}{0,0.6,0}

\definecolor{backcolour}{rgb}{0.95,0.95,0.92}
\definecolor{codegreen}{rgb}{0,0.6,0}

\lstdefinestyle{myStyle}{
    backgroundcolor=\color{white},   
    commentstyle=\color{codegreen},
    basicstyle=\ttfamily\footnotesize,
    breakindent=0em,
    breakatwhitespace=true,         
    breaklines=true,        
    columns=flexible,
    keepspaces=true,                 
    numbers=left,       
    numbersep=5pt,                  
    showspaces=false,                
    showstringspaces=false,
    showtabs=false,                  
    tabsize=2,
}
\makeatletter
\algnewcommand{\LineComment}[1]{\Statex \hskip\ALG@thistlm \(\triangleright\) #1}
\makeatother
\newcommand{\BibTeX}{B\kern-.05em{\sc i\kern-.025em b}\kern-.08em\TeX}

\newcommand{\sS}{\mathcal{S}}
\newcommand{\sA}{\mathcal{A}}
\newcommand{\sP}{\mathcal{P}}
\newcommand{\sR}{\mathcal{R}}

\begin{document}


\begin{frontmatter}


\paperid{642} 


\title{Evolutionary Reinforcement Learning via \\Cooperative Coevolution}

\author[A,B]{\fnms{Chengpeng}~\snm{Hu}}
\author[B,A]{\fnms{Jialin}~\snm{Liu}}
\author[C]{\fnms{Xin}~\snm{Yao}}

\address[A]{Guangdong Provincial Key Laboratory of Brain-inspired Intelligent Computation, Department of Computer Science and Engineering, Southern University of Science and Technology, Shenzhen, China}
\address[B]{Research Institute of Trustworthy Autonomous System, Southern University of Science and Technology,\\ Shenzhen, China}
\address[C]{Department of Computing and Decision Sciences, Lingnan University, Hong Kong SAR, China}




\begin{abstract}
Recently, evolutionary reinforcement learning has obtained much attention in various domains. Maintaining a population of actors, evolutionary reinforcement learning utilises the collected experiences to improve the behaviour policy through efficient exploration. 
However, the poor scalability of genetic operators limits the efficiency of optimising high-dimensional neural networks.
To address this issue, this paper proposes a novel cooperative coevolutionary reinforcement learning (CoERL) algorithm. Inspired by cooperative coevolution, CoERL periodically and adaptively decomposes the policy optimisation problem into multiple subproblems and evolves a population of neural networks for each of the subproblems. Instead of using genetic operators, CoERL directly searches for partial gradients to update the policy. Updating policy with partial gradients maintains consistency between the behaviour spaces of parents and offspring across generations.
The experiences collected by the population are then used to improve the entire policy, which enhances the sampling efficiency.
Experiments on six benchmark locomotion tasks demonstrate that CoERL outperforms seven state-of-the-art algorithms and baselines.
Ablation study verifies the unique contribution of CoERL's core ingredients.
\end{abstract}

\end{frontmatter}


\section{Introduction}

Reinforcement learning (RL) has made adequate advancements in various domains such as video games~\cite{mnih2015human,vinyals2019grandmaster}, Go~\cite{silver2016mastering} and robotic control~\cite{haarnoja2018soft} with a competitive level beyond human. 
Evolutionary reinforcement learning (ERL)~\cite{khadka2018evolution,bai2023evolutionary} combines evolutionary computation and RL to solve sequential decision-making problems, capitalising on the benefits of evolution in exploration.
ERL maintains a population of parameterised actors, called individuals, which interact with the environments to collect experiences.
Individuals with higher reward-based fitness are more likely to be selected as the parents, to which mutation and crossover operators are applied for reproduction. Another RL agent called \emph{learner}, learns from diverse experiences sampled by the maintained population of actors and substitutes the individual with the worst fitness in the population periodically, i.e., injecting its gradient information for better convergence.

However, ERL introduces a scalability problem, attributed to the genetic operators used for reproducing new individuals~\cite{yao1999evolving,salimans2017evolution,lehman2018safe}. Minor perturbations in the parameter space, caused by genetic operators, lead to significant divergences in behaviour space~\cite{lehman2018more,lehman2018safe}. Consequently, an offspring might not inherit its parents' behaviours by merely exchanging fragments of parameters. Thus, this inconsistency between parameter space and behaviour space does not align with the principles of evolution~\cite{yao1993review}.

\citet{lehman2018safe} proposed a safe mutation operator based on the sensitivity of neural networks' outputs, which slightly adjusts the parameters of neural networks to ensure consistency. State-based crossover~\cite{gangwani2018genetic}, as well as distillation crossover and proximal mutation operator~\cite{bodnar2020proximal}, improve the traditional genetic operators using backpropagation. However, those operators are gradient-based, which introduce in-negligible extra computational costs.
Nevertheless, the works of \cite{nesterov2017random} and \cite{yang2022evolutionary} directly searched for the parameters of neural networks using the cooperative coevolution, in which the policy optimisation problem is decomposed into multiple low-dimensional subproblems.
However, the experiences of optimising subproblems are not fully utilised, resulting in inefficient training~\cite{khadka2018evolution}.

To address the scalability issue, we propose a novel cooperative coevolutionary reinforcement learning (CoERL) algorithm. The optimisation of a high-dimensional neural network is decomposed into multiple low-dimensional subproblems periodically and adaptively by cooperative coevolution. A population is resampled for each subproblem, in which each individual maintains the same neural network architecture.
CoERL directly searches for partial gradients and updates the entire policy.
The partial gradient searched via the population guides a proximal update on the subproblem, diminishing inconsistency between the behaviour spaces of parents and offspring.
Experiences collected during evolution are then used to enhance RL for better sampling efficiency.

Our main contributions are summarised as follows:
\begin{itemize}
    \item This paper proposes CoERL, a novel cooperative coevolutionary reinforcement learning algorithm to address the scalability issue of ERL, which also improves sample efficiency during training. 
    \item This paper proposes a partially updating strategy for policy improvement based on cooperative coevolution, which costs less and maintains the consistency between the behaviour spaces of parents and offspring across the evolution.
    \item Experimental results on six benchmark locomotion tasks demonstrate the comparable performance of CoERL, compared with seven state-of-the-art algorithms and baselines. Ablation study shows the unique contributions of CoERL's core ingredients.
\end{itemize}

\section{Background}
\label{sec:back}

In this section, we introduce some preliminary, cooperative coevolution and recent progress of ERL.

\subsection{Preliminary: Markov decision process}
Markov decision process (MDP)~\cite{sutton2018reinforcement}
is defined as a tuple $(\sS,\sA,\sR,\sP,\gamma)$, where $\sS$ is the set of states, $\sA$ is the set of actions, $\sR:\sS\times \sA \times \sS \mapsto \mathbb{R}$ is the reward function, $\sP:\sS\times \sA \times \sS \mapsto [0,1]$ is the transition probability function, and $\gamma$ is the discount factor. 
A policy $\pi$ is a mapping from states to probability distributions for acting, where $\pi(a_t|s_t)$ is the probability of taking action $a_t$ in state $s_t$ at time $t$. 
The goal of the MDP is to optimise a parameterised policy $\pi_\theta$ that maximises the discounted cumulative reward, formulated as:
\begin{equation}
\max_{\theta}~J(\theta)=\mathbb{E}_{\tau \sim\pi_\theta}[\sum_{t=0}^{\infty} \gamma^t \sR(s_t,a_t,s_{t+1})],
\end{equation}
 where $s_0$ is an initial state. $\tau \sim\pi_{\theta}$ denotes a trajectory $(s_0,a_0,s_1,a_1,\dots, s_t,a_t,s_{t+1})$ sampled from $\pi_{\theta}$.

\subsection{Cooperative coevolution}
Cooperative coevolution simulates the interactions of species living in an ecosphere~\cite{ma2018survey}.
~\citet{potter1994cooperative} proposed the first  cooperative coevolution (CC) algorithm, called cooperative coevolutionary genetic algorithm (CCGA). CCGA decomposes an optimisation problem into multiple subproblems. A population is initialised for each subproblem with a global context vector for fitness evaluations. 
For example, given an optimisation problem with $n$ variables, the fitness of a partial solution $x_{ij}$, i.e., the $j$-th individual in the population of $i$-th subproblem, can be evaluated by directly replacing the variable $b_j$ in the context vector $\bm{b}$. Each variable in $\bm{b}$ is then updated with the best individual for the corresponding subproblem.

Problem decomposition referring to the grouping of variables is critical in CC~\cite{ma2018survey}. Incorrect grouping leads to a local optimum, which is limited by the decomposed problem itself~\cite{panait2010theoretical,yang2017turning}. 
Intuitively, static grouping and random grouping~\cite{potter1994cooperative} are easy to implement and require no additional computational budget or knowledge. However, the effectiveness of the general strategies is unclear and requires some presets. Interaction-based grouping~\cite{liu2013scaling} and landscape-aware grouping~\cite{wu2022cooperative} decompose the problem according to the characteristics of variables such as correlation and landscape. However, additional costs for fitness evaluations are required.

Another issue concerns the evaluation of individuals within a subproblem's population, also known as the credit assignment problem, since individuals in a population represent only part of the entire solution. Collaboration for fitness evaluation with other subproblems' populations is required.
A global context vector can be constructed by collaborators (i.e., individuals) selected from populations of subproblems. 
Popular selection methods, such as single best collaborator~\cite{potter1994cooperative}, elite collaborator~\cite{glorieux2015improved}, and random collaborator selection, are typically chosen based on the specific domain~\cite{ma2018survey}.

Cooperative coevolution, served as a useful technique for solving high-dimensional black-box optimisation problems, is expected to benefit the policy optimisation. 
In this paper, we apply random grouping and single best collaboration for low cost and simplicity.

\subsection{Evolutionary reinforcement learning}
Evolutionary algorithms (EAs) have been successfully applied to solve RL problems, where neuroevolution and evolution-guided policy gradient are two main trends in this area~\cite{moriarty1999evolutionary,yao1999evolving,whiteson2012evolutionary,bai2023evolutionary}.

\subsubsection{Neuroevolution}

Neuroevolution directly optimises the parameters or architectures of neural networks without considering the underlying mechanism like gradients, making it suitable for evolving RL policy and tackling real-world problems~\cite{liu2014meta,cauwet2016algorithm,liu2020self,galvan2021neuroevolution,Yangp@2023reducing}. 
Fitness metrics formulated on rewards are used to select good individuals for reproducing the next generation by mutation and crossover. Hence, neuroevolution can intuitively handle complicated rewards such as non-convex and non-differentiable cases~\cite{salimans2017evolution,cauwet2016algorithm}.
Population-based training also encourages exploration. 
However, \citet{nesterov2017random} indicated that EAs scale poorly with the increase of parameters.

\citet{gong2020evolving} combined coevolution and backpropagation to solve classification tasks, while the framework is limited by traditional reproduction and specific domains.
\citet{yang2022evolutionary} proposed a CC framework based on negatively correlated search to tackle this issue, which directly searches for the parameters of a neural network. However, it only utilises the final accumulative reward as fitness, resulting in a waste of experiences collected during the evolution.

\begin{figure}[t]
    \centering
    \includegraphics[width=0.83\linewidth]{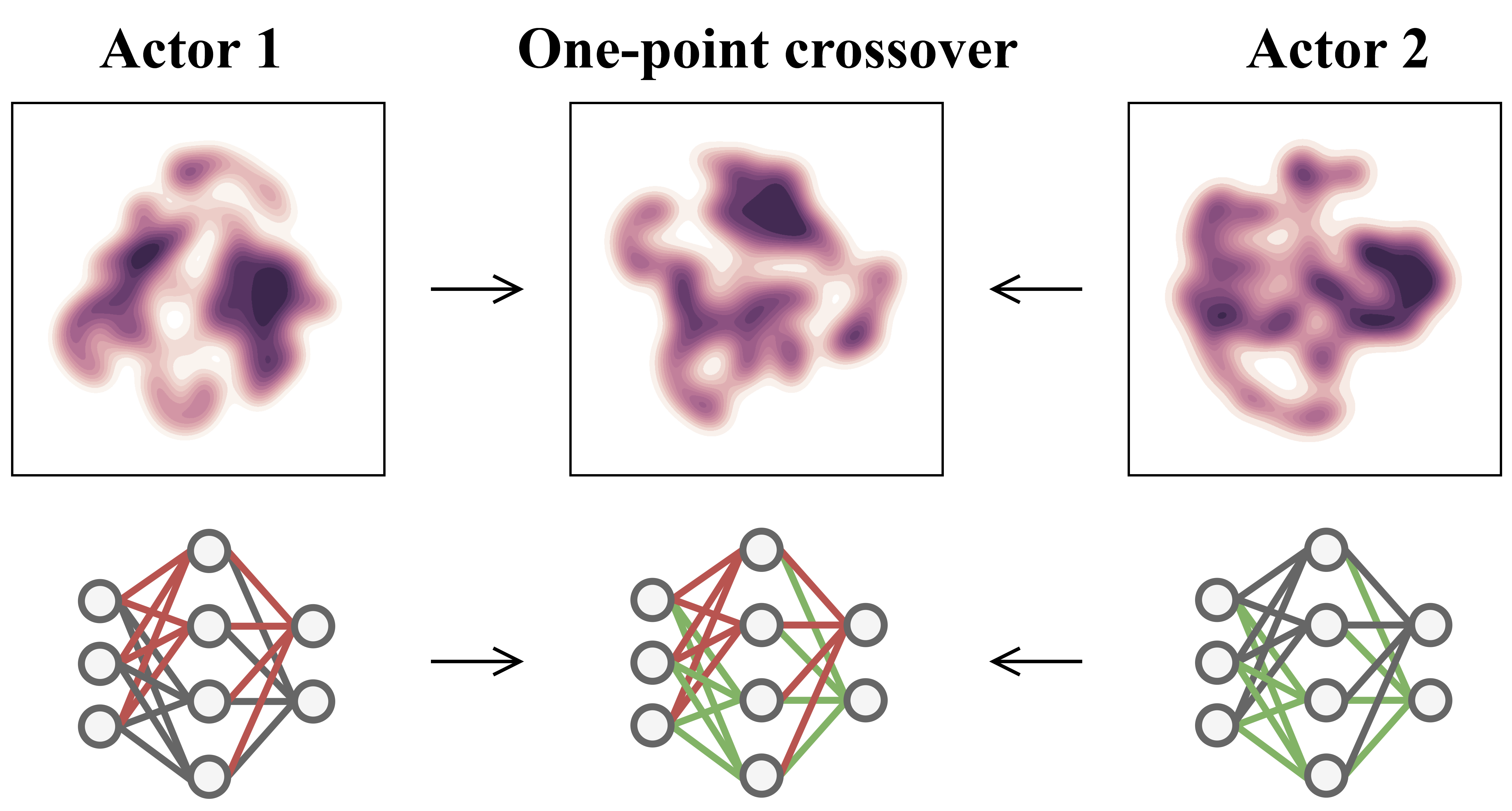}
    \caption{Inconsistency phenomenon between the behaviour spaces of parents and offspring by applying one-point crossover to exchange partially the parameters of Actor 1 (red) and Actor 2 (green). Subfigures show the feature maps of behaviour spaces decompressed by t-SNE~\cite{van2008visualizing}.
    }
    \vspace{5mm}
    \label{fig:onepointc}
\end{figure}

Besides, traditional genetic operators limit EA's capability of addressing large-scale optimisation problems. 
The direct operations on the parameter space lead to an inconsistency between the behaviour spaces of parents and offspring, where parental behaviours may be typically forgotten~\cite {lehman2018safe,gangwani2018genetic,bodnar2020proximal}.
Figure \ref{fig:onepointc} shows an example of applying the one-point crossover, in which behaviour spaces of actors are decompressed by t-SNE~\cite{van2008visualizing}.
It is easy to see from Figure \ref{fig:onepointc} that directly exchanging the parameter fragments of the parents leads to a forgetting phenomenon in the behaviour space of the offspring.

\subsubsection{Evolution-guided policy gradient
}
\begin{figure*}[!h]
    \centering
    \includegraphics[width=0.87\linewidth]{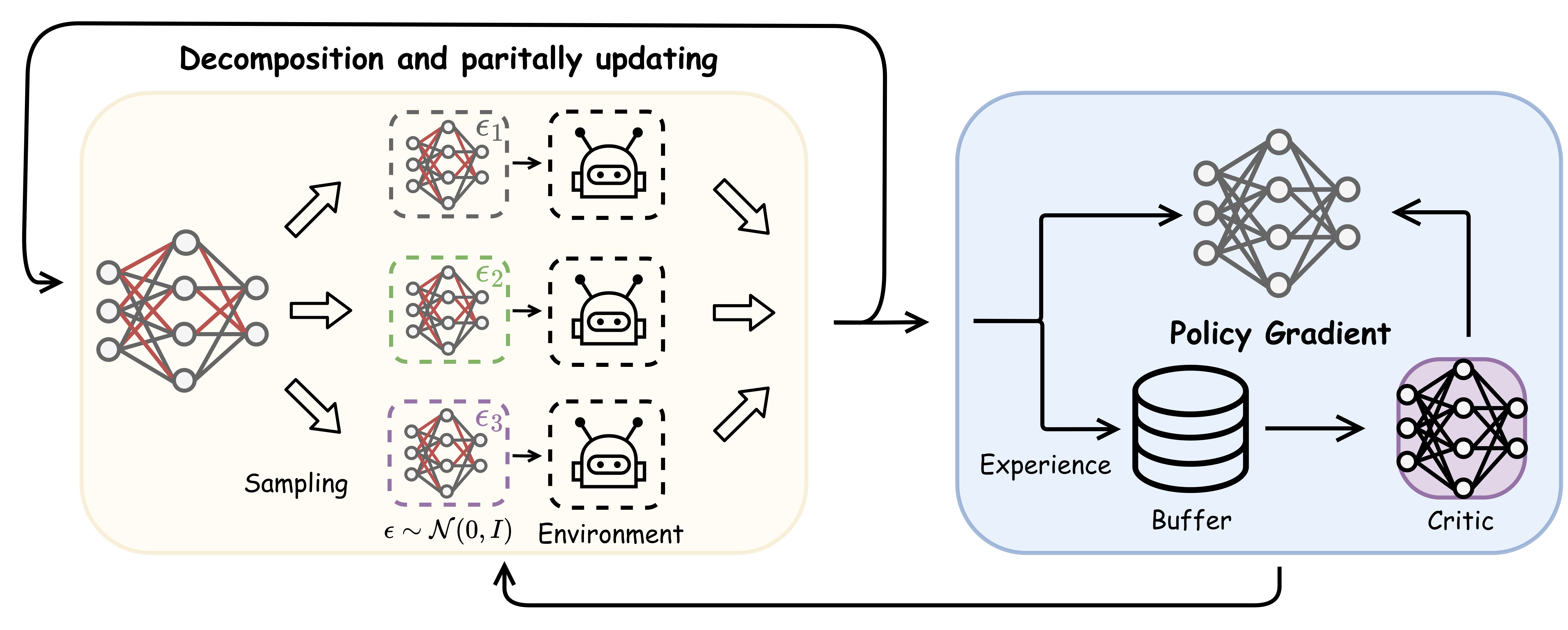}
    \caption{Diagram of CoERL. The CoERL algorithm begins by decomposing the policy optimisation problem parameterised by a neural network into multiple subproblems and searching for partial gradients to update the policy. Subsequently, the explored experiences are gathered to further refine the policy outcome using the MDP-based RL.}
    \label{fig:ccoerl}
    \vspace{5mm}
\end{figure*}
Evolution-guided policy gradient combines neuroevolution and MDP-based RL~\cite{khadka2018evolution}. In the EA loop, actors, also called individuals, interact with the environments and are evolved by genetic operators. Experiences collected during the evaluations, i.e., transitions, are stored in a replay buffer and leveraged in the RL loop. \citet{khadka2019collaborative} proposed a portfolio of policies and dynamically allocated computational resources to train different policies, extending ERL.
CEM-RL~\cite{pourchot2018cemrl} leveraged cross-entropy method (CEM) to delayed deep deterministic policy gradient (TD3). 
Without using genetic operators, CEM-RL samples actors from the policy distribution estimated in previous generations. \citet{gangwani2018genetic} applied imitation learning to the crossover operator, which use trajectories of two parents to train an offspring.
\citet{bodnar2020proximal} improved the mutation and crossover by using local replay memories and the critic of the RL learner, namely proximal distilled evolutionary reinforcement learning (PDERL). 
New offsprings are produced by exchanging the local replay buffer or perturbation according to the sensitivity of the actions in PDERL~\cite{bodnar2020proximal}.
Genetic algorithm has also been combined with RL in the work of~\cite{marchesini2020genetic}, where individuals are generated by the noisy mutation only. 
To tackle the computationally expensive evaluation, \citet{Wang2022surrogate} extended the surrogate model to ERL, which estimates the fitness of each individual based on the approximated value function. \citet{zhu2024two} addressed the exploration and exploitation issue by maintaining actor and critic populations.

Although ERL inherits the advantages of neuroevolution, enabling it to address the temporal credit assignment problem with sparse rewards, it faces the challenge of poor scalability with EAs. Recent efforts have primarily concentrated on enhancing genetic operators, but at the expense of increased computational costs.

\section{ Cooperative
coevolutionary reinforcement learning}
\label{sec:coerl}

Aiming at addressing the poor scalability of ERL, we propose a novel cooperative coevolutionary reinforcement learning (CoERL) algorithm~\footnote{Code: \url{https://github.com/HcPlu/CoERL}}.  
Pseudo-code and diagram of CoERL are provided in Algorithm \ref{alg:ccerl} and Figure \ref{fig:ccoerl}, respectively.

\input{codes/ccerl}

\subsection{Collaboration between cooperative coevolution and reinforcement learning}
CoERL maintains collaborative loops between CC and RL.
In the CC loop, a neural network (a policy) is converted into a position-fixed vector with real numbers. The policy optimisation problem, is divided into multiple subproblems, in which parameters of the neural network are grouped.
The decomposed subproblems have no intersection, while their union collectively constitute the entire parameter space. 
Given a policy $\pi_{\theta}$ parameterised by $\theta$, it ensures that $\theta^{\langle\mathcal{I}_i\rangle}\cap\theta^{\langle\mathcal{I}_j\rangle}=\varnothing,\forall i\neq j, 1\leq i \leq m, 1 \leq j \leq m$ and $\theta^{\langle\mathcal{I}_1\rangle}\cup\theta^{\langle\mathcal{I}_2\rangle}\cup\cdots\cup\theta^{\langle\mathcal{I}_m\rangle}=\theta$, where $m$ is the number of subproblems and $\langle\mathcal{I}_i\rangle$ denotes the parameter indices of subproblem $j$.
To optimise a subproblem, a population is maintained and generated using perturbation methods instead of traditional genetic operators. Practically, each individual in the population shares the same neural network structure with the freezing quotient set of the subproblem. Only the variables in the subproblem, which form part of the neural network, are perturbed by the noises from the distribution.

Individuals consistently interact with the environment. The cumulative reward obtained by the individual is treated as its fitness. 
Then, noised individuals and their corresponding fitness are used to update the entire policy.
Notably, subproblems are not optimised separately. The policy optimised according to the preceding subproblem is regarded as the complementary base of the subsequent subproblem. Thus, connections between different subproblems are built.

After updating the policy within each subproblem, CoERL proceeds to optimise the policy via MDP-based RL. Experiences such as transitions produced by the individuals during the cooperative coevolution loop are collected in a replay buffer. Batches sampled from the replay buffer are then used to update the policy via policy gradient, which fully utilises the temporal information. Sections~\ref{sec:cc} and~\ref{sec:rl} detail the two loops, respectively.

\subsection{Partially updating via cooperative coevolution
}
\label{sec:cc}

Given a policy $\pi_{\theta}$ parameterised by $\theta$, CoERL decomposes the policy optimisation problem into $m$ subproblems in the CC loop. 
The subproblem $j$ of policy optimisation, $\theta^{\langle\mathcal{I}_j\rangle}$ indexed by parameters indices $\langle\mathcal{I}_j\rangle$, and the quotient $\hat{\theta}^{\langle\mathcal{I}_j\rangle} = \theta/\theta^{\langle\mathcal{I}_j\rangle}$ constitute the entire parameter space, i.e., $\theta=\left[\theta^{\langle\mathcal{I}_j\rangle}:\hat{\theta}^{\langle\mathcal{I}_j\rangle}\right]$. Then, for each subproblem $j$, a corresponding actor $\pi_{\theta^{\langle\mathcal{I}_j\rangle}}$ is constructed by freezing the quotient part.
CoERL partially updates the policy by iteratively optimising each subproblem. A population is first sampled via a specific distribution $P^{\langle\mathcal{I}_j\rangle}$. Each individual is an actor $\pi_{\theta^{\langle\mathcal{I}_j\rangle}}$, where $\theta^{\langle\mathcal{I}_j\rangle} \sim P^{\langle\mathcal{I}_j\rangle}$.
All individuals in the population of the subproblem share the same quotient part $\hat{\theta}^{\langle\mathcal{I}_j\rangle}$. Then, the fitness of each individual $f(\pi_{\theta^{\langle\mathcal{I}_j\rangle}})$ is determined by the cumulative reward. The expected fitness under an arbitrary distribution is formulated as Eq. \eqref{eq:gen_j}:
\begin{equation}
    J(\theta^{\langle\mathcal{I}_j\rangle}) = \mathbb{E}_{\theta^{\langle\mathcal{I}_j\rangle}}[f(\pi_{\theta^{\langle\mathcal{I}_j\rangle}})]\\ 
    = \int f(\pi_{\theta^{\langle\mathcal{I}_j\rangle}})p(\theta^{\langle\mathcal{I}_j\rangle})d\theta^{\langle\mathcal{I}_j\rangle},\label{eq:gen_j}
\end{equation}
where $p(\theta^{\langle\mathcal{I}_j\rangle})$ denotes the density. Then, we can write the gradient form using the ``log-likelihood trick". The estimation of the gradient by maintaining a population with size $\mu$ is shown as follows:
\begin{equation}
      \nabla_{\theta^{\langle\mathcal{I}_j\rangle}} J(\theta^{\langle\mathcal{I}_j\rangle}) = \mathbb{E}_{\theta^{\langle\mathcal{I}_j\rangle}}\left[f(\pi_{\theta^{\langle\mathcal{I}_j\rangle}})\nabla_{\theta}\log(p(\theta^{\langle\mathcal{I}_j\rangle}))\right].
\end{equation}

\begin{figure}[!h]
    \centering
    \includegraphics[width=0.86\linewidth]{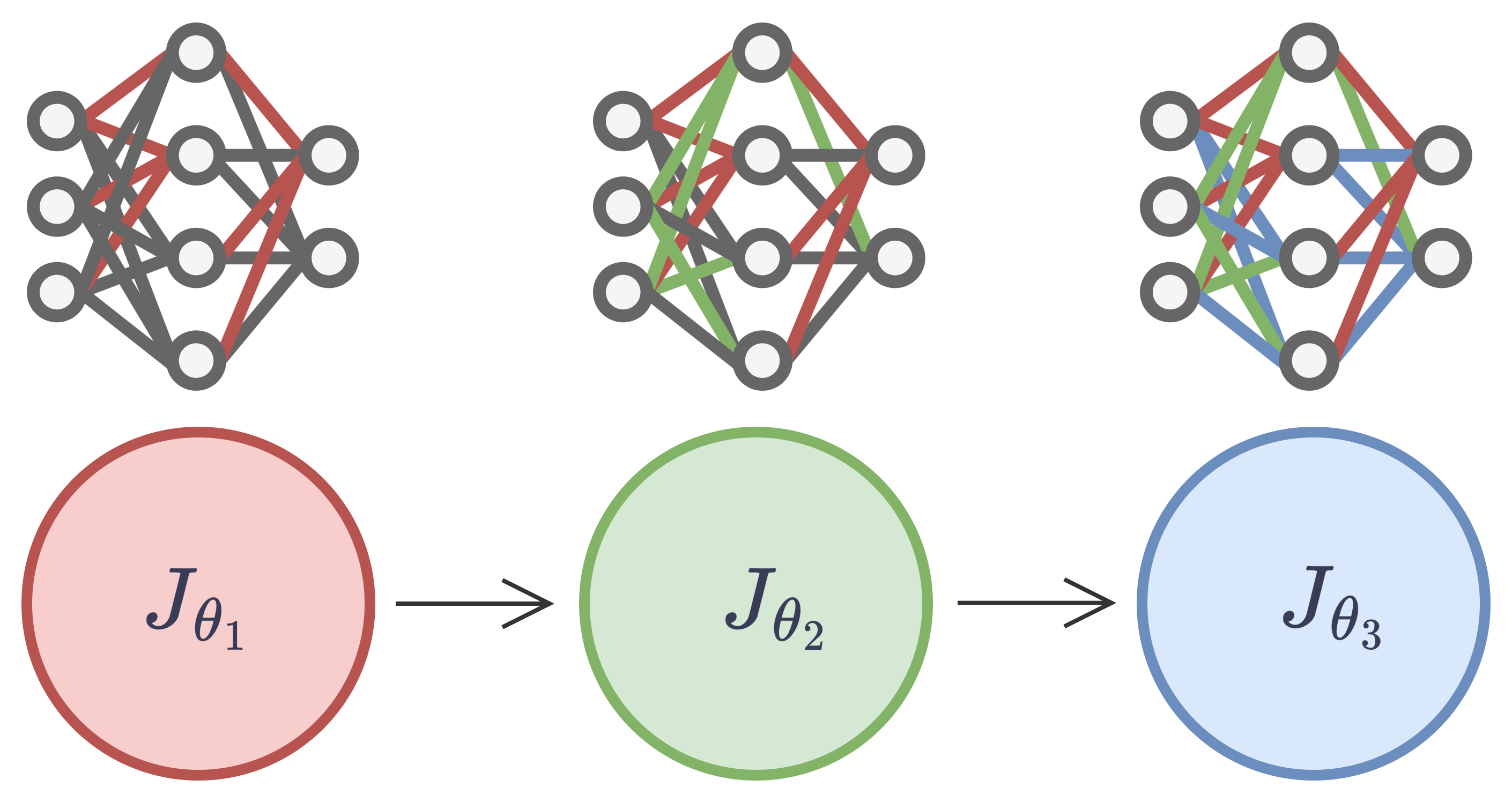}
    \caption{An example of partial updating via CC. Three subproblems are highlighted in red, green, and blue, respectively. For each subproblem, the policy inherits the partial gradient from the previous subproblem. Eventually, all parameters are updated once and only once.}
    \vspace{5mm}
    \label{fig:pg3}
\end{figure}

In the case of factored Gaussian distribution with the deviation $\sigma$,
we can set $\theta^{\langle\mathcal{I}_j\rangle}$ as the mean vector. Then, the expected fitness of Eq.~\eqref{eq:gen_j} is rewritten as follows:
\begin{equation}
    \mathbb{E}_{\theta^{\langle\mathcal{I}_j\rangle}}[f(\pi_{\theta^{\langle\mathcal{I}_j\rangle}})] = \mathbb{E}_{\epsilon\sim \mathcal{N}^{\langle\mathcal{I}_j\rangle}(0,I)}[f(\pi_{\theta^{\langle\mathcal{I}_j\rangle}+\sigma\epsilon})].\label{eq:gau_mean}
\end{equation}
Practically, each individual, i.e., $\pi_{\psi_i}$, the actor in the population is produced by a perturbation operation $\psi_i = \theta^{\langle\mathcal{I}_j\rangle}+\epsilon_i$, where $ \epsilon_i\sim \mathcal{N}^{\langle\mathcal{I}_j\rangle}(0,I)$.
The average estimated gradient in this case according to Eq.~\eqref{eq:gau_mean} is shown as follows:
\begin{equation}
    \nabla_{\theta^{\langle\mathcal{I}_j\rangle}}\mathbb{E}_{\theta^{\langle\mathcal{I}_j\rangle}}[f(\pi_{\theta^{\langle\mathcal{I}_j\rangle}})] \cong \frac{1}{\mu\sigma}\sum^{\mu}_{i=1}[f(\pi_{\theta^{\langle\mathcal{I}_j\rangle}+\sigma\epsilon_i})\epsilon_i], \label{eq:e_p_g}
\end{equation}
where $\mu$ is the population size.
The estimated partial gradient enables searching for a good optimisation direction for each subproblem. Instead of independently optimising each subproblem, we introduce the concept of coevolution by the divide-and-conquer strategy, in which each subproblem is optimised iteratively. 
Using Figure \ref{fig:pg3} as an example, a neural network optimisation problem (weights) is decomposed into 3 subproblems (3 subsets), highlighted by red, green and blue, respectively. Each subproblem aims at optimising a subset of the weights. First, weights in red will be updated, while those in black will be frozen. Then, weights in green will be updated, while both weights in red and black will be frozen. Similar steps proceed until all weights are updated once.
Optimising subproblems is not independent but in a cascade way. Optimisation of the current subproblem relies on the previous subproblem.

The idea of coevolution builds the bridge between subproblems. The estimated gradient of the preceding subproblem can be regarded as momentum, which facilitates the subsequent optimisation. Additionally, the time complexity of the partially updating is $\mathcal{O}(|\theta|)$ with linearly scaling the number of parameters. This complexity is easy to get since each subproblem is the complimentary set of other subproblems. This complexity is much smaller than that of current gradient-based operators such as distilled crossover~\cite{bodnar2020proximal}, which requires backpropagation with batches.

\def\nose{
\begin{equation}
  \nabla_{\psi} \mathbb{E}_{\theta\sim p_{\psi}}F(\theta) = \mathbb{E}_{\theta\sim p_{\psi}}F(\theta)\nabla_{\psi}\log(p_{\psi}(\theta)),
\end{equation}

\begin{equation}
  \nabla_{\psi} \mathbb{E}_{\textbf{w}_i\sim p_{\psi}}F(\left[\theta_i:\textbf{w}_i\right]) = \mathbb{E}_{\textbf{w}_i\sim p_{\psi}}F(\left[\theta_i:\textbf{w}_i\right])\nabla_{\psi}\log(p_{\psi}(\textbf{w}_i)),
\end{equation}
}

\subsection{Leveraging temporal information}
\label{sec:rl}

To fully utilise the collected experiences, we choose soft actor-critic (SAC)~\cite{haarnoja2018soft} as the base algorithm in the RL loop, which optimises the policy in an off-policy way. Besides, the actor-critic architecture allows us to train the actor and critic separately. Hence, a population can be directly generated from an actor. The optimised actor after the CC loop is then used in policy improvement directly.
SAC changes the objective of RL by adding the entropy term. The entropy of the policy is regarded as an extra rewarding signal. Larger entropy indicates a greater tendency for exploration.   
The $Q$ function is parameterised by $\phi$, and the target $Q$ value using reparameterisation $\tilde{a}'\sim \pi_{\theta}(\cdot|s')$ is shown as follows:
\begin{equation}
    y(r,s') = r+\gamma(\min_{j=1,2}Q_{\phi_j}(s',\tilde{a}')-\alpha_s\log{\pi_{\theta}(\tilde{a}'|s')}),
\end{equation}
where $\alpha_s$ is the temperature coefficient. If $s'$ is the end of the episode, then $y(r,s')=r$.
The loss function of the Q-network is shown as follows:
\begin{equation}
    L(\phi_i) = \mathop{\mathbb{E}}\limits_{(s,a,r,s')}\left[(Q_{\phi_i}(s,a)-y(r,s'))^2\right].
\end{equation}
According to the additional entropy term, the objective of SAC is
\begin{equation}
    \max_{\theta}~J(\theta) = \mathop{\mathbb{E}}\left[\min_{j=1,2}Q_{\phi_j}(s,\tilde{a}_{\theta})-\alpha_s\log{\pi_{\theta}(\tilde{a}_{\theta}|s)}\right],
\end{equation}
where $\tilde{a}_{\theta}$ is sampled via the reparameterisation trick.

\section{Experiments}
\label{sec:exp}
\begin{figure*}[t]
\centering

    \subfigure{
           \includegraphics[width=.321\linewidth]{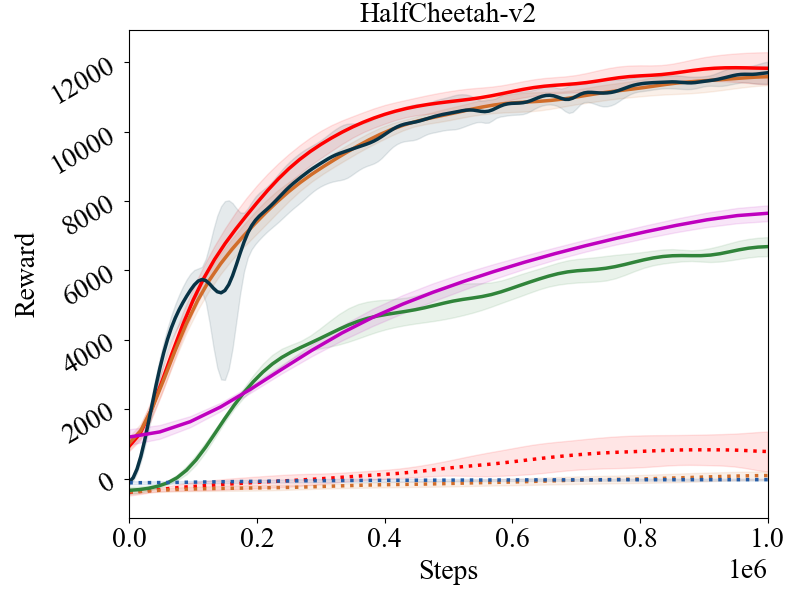} 
    
    }
    \subfigure{
     \includegraphics[width=.321\linewidth]{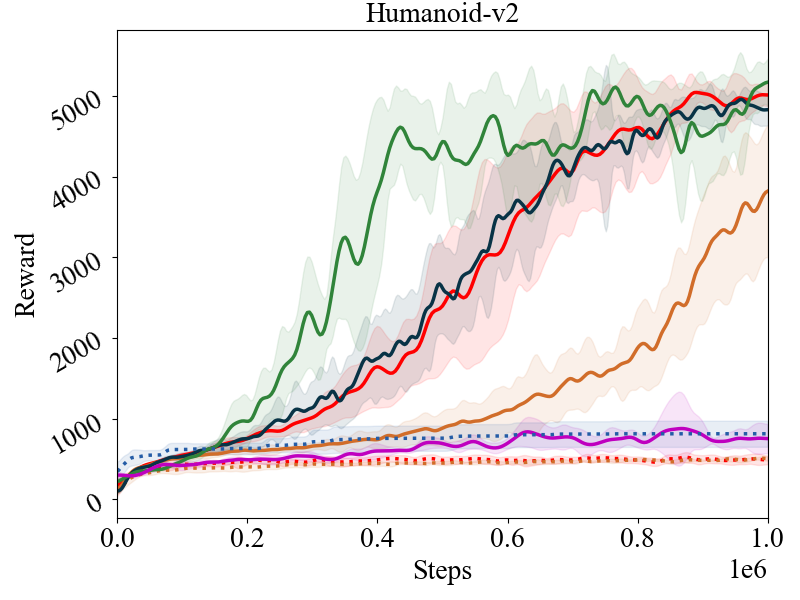}
    }
\subfigure{         \includegraphics[width=.321\linewidth]{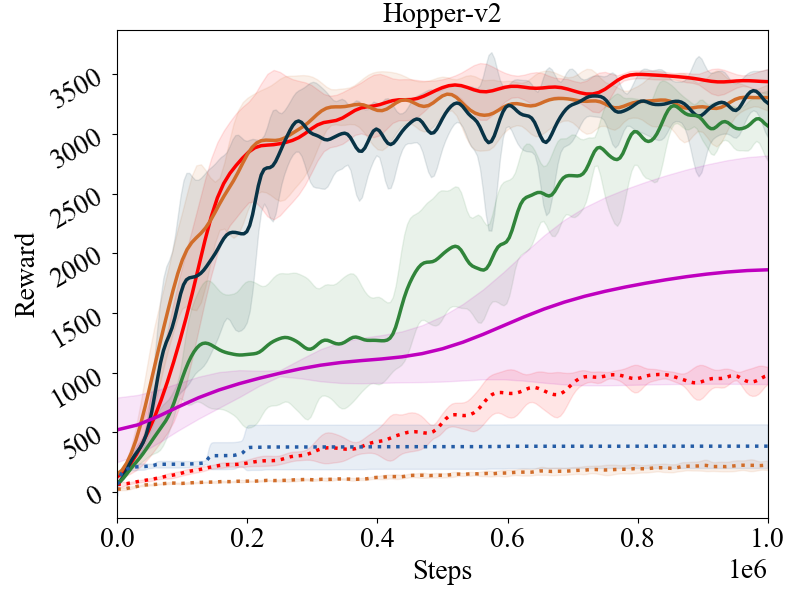}}

\subfigure{           \includegraphics[width=.321\linewidth]{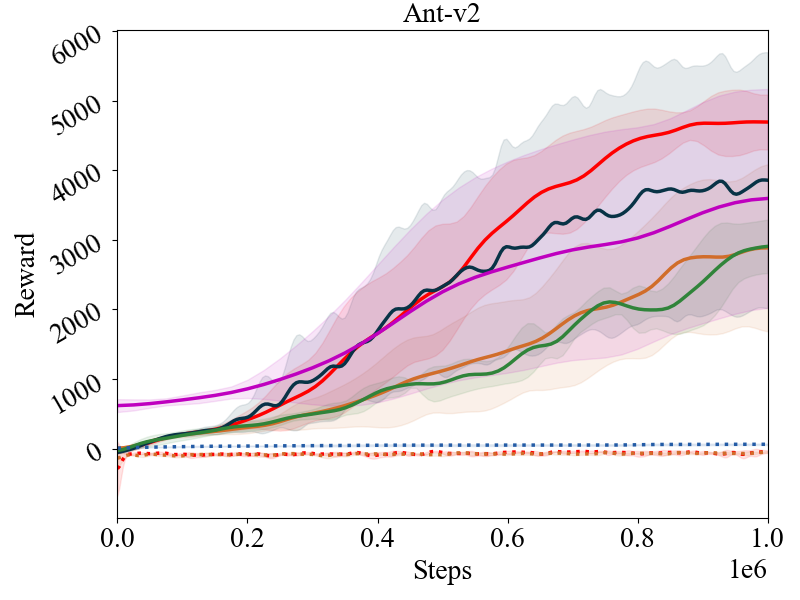} }
\subfigure{         \includegraphics[width=.321\linewidth]{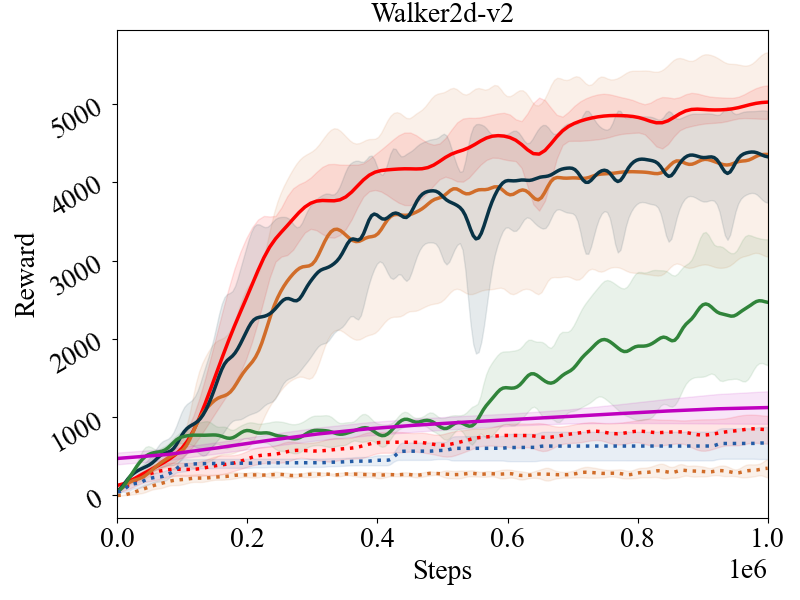}}
\subfigure{         \includegraphics[width=.321\linewidth]{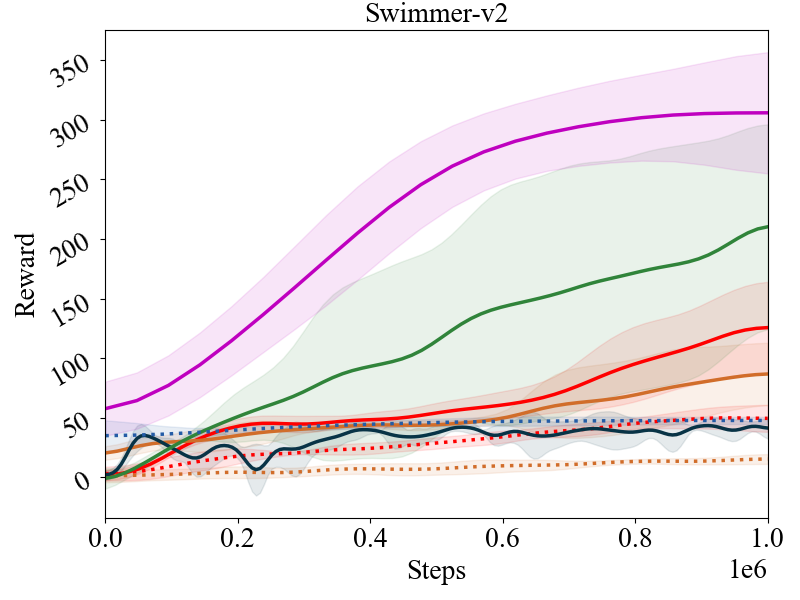}}
\hspace{30mm}
\subfigure{         \includegraphics[width=0.85\linewidth]{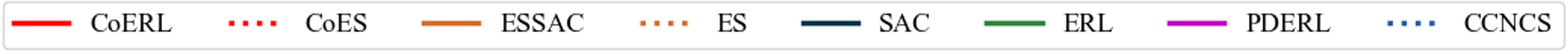}}
    \caption{Training curves on six locomotion tasks. Each algorithm is trained for 1e6 timesteps with five different random seeds.}
    \label{fig:training}
    \vspace{5mm}
\end{figure*}

CoERL is compared with four advanced RL and ERL algorithms, namely SAC~\cite{haarnoja2018soft}, ERL~\cite{khadka2018evolution}, PDERL~\cite{bodnar2020proximal}, and CCNCS~\cite{yang2022evolutionary} on six benchmark locomotion tasks, including Ant-v2, HalfCheetah-v2, Humanoid-v2, Hopper-v2, Walker2d-v2, and Swimmer-v2 from Mujoco~\cite{todorov2012mujoco} integrated in the OpenAI gym~\cite{OpenAIGym}. Additionally, CoERL is split into three independent methods for conducting ablation study, namely cooperative coevolutionary evolution strategies (CoES), evolution strategies with soft actor-critic (ESSAC), and evolution strategies (ES)~\cite{salimans2017evolution}.

\subsection{Settings}
\begin{table}[b]
    \centering
        \caption{Hyperparameters for all compared algorithms. ``-" denotes that the parameter is not involved.}
        \vspace{5mm}
        \begin{tabular}{l|c|c}
    \toprule
        \multirow{2}{*}{Algorithm} & \multirow{2}{*}{Pop size $\mu$} & Learning rate   \\
         & & Actor/Critic/Evolution  \\
        \midrule
        CoERL (ours)&6& 1e-3/ 1e-3/ 1e-3\\
CoES& 6 &-/-/ 1e-3\\
ESSAC& 6 &1e-3/ 1e-3/ 1e-3\\
ES~(\citet{salimans2017evolution})& 6& -/-/ 1e-3 \\
SAC~(\citet{haarnoja2018soft})&-&1e-3/ 1e-3/- \\
ERL~(\citet{khadka2018evolution}) &10 &3e-4/ 1e-4/-\\
PDERL~(\citet{bodnar2020proximal})& 10 & 5e-4/ 5e-3/-\\
CCNCS~(\citet{yang2022evolutionary}) &6 &-/-/-\\
\bottomrule
    \end{tabular}
    \vspace{5mm}
    \label{tab:hp}
\end{table}
CoERL is implemented with the Tianshou framework~\cite{weng2022tianshou}. The network structure is a fully connected neural network with a $\langle256,256\rangle$ linear layer.
$\gamma$ is 0.99. 
In this paper, we apply random grouping~\cite{yang2008multilevel,yang2022evolutionary} and single best collaboration~\cite{potter1994cooperative} for low cost and simplicity. The grouping number at each generation is randomly chosen from two, three and four, as~\citet{yang2022evolutionary} suggested. Hyperparameters of SAC, ERL, PDERL and CCNCS follow their original settings in \cite{haarnoja2018soft,khadka2018evolution,bodnar2020proximal,yang2022evolutionary}, respectively. Main hyperparameters are shown in Table~\ref{tab:hp}.
All algorithms are trained by 1e6 timesteps with five different random seeds.
CoERL's hyperparameters are arbitrarily set, and remain the same values for all tasks.
\subsection{Comparison results}
Figure \ref{fig:training} shows the training curves assembled with five different random seeds. Our CoERL, highlighted with red colour, presents comparable performance and superior convergence on \emph{HalfCheetah-v2}, \emph{Hopper-v2}, \emph{Ant-v2} and \emph{Walker2d-v2}.
Table \ref{tab:mujoco_5seeds} presents the final reward performance of CoERL and all comprised algorithms across six locomotion tasks. 
CoERL achieves the best average rewards on four tasks including \emph{HalfCheetah-v2}, \emph{Hopper-v2}, \emph{Ant-v2} and \emph{Walker2d-v2}. For example, in \emph{Ant-v2}, CoERL obtains the best average reward of 5037.22, while the second-best algorithm SAC only gets 3654.16. 
Besides, CoERL achieves the highest average rank of 1.67, considering all six tasks. CoERL decomposes the neural network and searches for the partial gradient for each subproblem. Each parameter of the entire neural network is updated only once within an iteration, with a complexity of $\mathcal{O}(|\theta|)$. The partial gradient estimation does not rely on the delicate calculation of batches from the dataset but only requires the final outcome of evaluations, which can be easily sped up on CPUs according to Eq. \eqref{eq:e_p_g}.

\input{tables/mujoco_5seeds_formatted}

We also notice that CoERL does not achieve outstanding performance in \emph{Swimmer-v2}, although its performance has been improved from 55.89 to 94.87 and 128.90 with the help of CC and the utility of temporal information by ESSAC. The poor performance of SAC might explain this case, as it only achieves 37.95, surpassing only the 15.91 achieved by ES. Given that CoERL is built upon SAC, it is not surprising that CoERL encounters the same local optimum as SAC, despite ultimately reaching 128.90.
PDERL, the improved version of ERL only get 765.55 in \emph{Humanoid-v2}, while our CoERL gets an average reward of 4642.05 and ERL gets 4677.19. It is attributed to the high dimensions of \emph{Humanoid-v2} for the peculiar performance of PDERL. The observation dimension of \emph{Humanoid-v2} is 376, which is the largest among six tasks. The distilled crossover maintains the consistency of behaviour spaces while exchanging parameters using supervised learning techniques. However, this improvement leads to higher computational demands during backpropagation. Our CoERL, in contrast, searches for the partial gradient at a lower computational cost than backpropagation, while still preserving consistency.

Moreover, none of the pure evolution-based algorithms, such as ES~\cite{salimans2017evolution} and CCNCS~\cite{yang2022evolutionary}, demonstrates overall promising performance, even though CoES and CCNCS achieve superior average rewards than SAC in \emph{Swimmer-v2} with 55.89 and 47.71, respectively. As discussed previously, evolution-based algorithms have long struggled with the scalability issue~\cite{ma2018survey}. And the temporal experiences are barely utilised in evolution, resulting in inefficient training. Therefore, more computational time is required to converge. Our CoERL overcomes those issues and achieves the highest rank with 1.67.

\subsection{Ablation study}
To fully verify the effectiveness of CoERL, we examine the contribution of its components independently. The evolution part is split into coevolution strategies (CoES) and evolution strategies (ES). Regarding the RL part, we consider pure RL algorithm, SAC, and the simplified version without coevolution, ESSAC.
The performance of CoERL and its ingredients, including CoES, ES, SAC, and ESSAC, are shown in Table \ref{tab:mujoco_5seeds} and Figure \ref{fig:training}. It is evident from the results that none of the ablated algorithms like ES, CoES and ESSAC outperforms CoERL, indicating their unique contributions to CoERL.

Evolution-based algorithms such as CoES and ES only utilise the outcome of the evaluation as the fitness. Although EAs show some advantages in tackling the credit assignment problem with sparse reward~\cite{salimans2017evolution}, it wastes adequate temporal information collected during evaluation. On the other hand, modern MDP-based RL algorithms tend to use temporal information, i.e., transition-based experiences, while ignoring the long-term cumulative reward. This dilemma is similar to the trade-off between Monte-Carlo sampling and temporal difference~\cite{sutton2018reinforcement}, which seeks to balance between bias and variance of the estimated value function. In our case, the usage of temporal information becomes an issue as traditional evolution-based algorithms merely provide a solution to utilise it.

Instead of a single evolution loop, CoERL maintains an additional MDP-based RL loop for better use of temporal information. 
The RL loop reuses the experiences collected by the actors in the evolution. Since the evolution loop still inherits the surviving technique based on fitness, a variation between the target policy and the behaviour policy, i.e., individuals, is introduced. So the choice of MDP-based RL has to be the off-policy version~\cite{khadka2018evolution}. 

Intuitively, the additional RL loop proceeds to optimise the policy, which achieves an efficient sample utility for picking up the wasted experiences. The policy is improved using value approximation in a fine-grained way.
At the same time, when looking into the way of collecting experiences, we find that the underlying mechanism of the reproduction shows an advantage on exploration. New individuals are generated through proximal perturbation in CoERL, ensuring a diverse range of experiences.

Besides, decompostion is the essential part of CoERL. The simple random grouping is applied to decompose the policy in CoERL. The unique performance is evaluated via the ablation study. It's possible to improve this part by using other decomposition techniques like~\cite{wu2022cooperative}, which learns the problem structure and variable dependency for decomposition. However, it might be computationally expensive. Another possible way is to fix the number of subproblems. As the neural network's structure is known, we can decompose it with different layers. Of course, determining the dependency of layers and merging different layers as a subproblem are crucial.

\begin{table*}[t]
\caption{Average running time (minutes) of algorithms over five trials with different random seeds.}
\vspace{3mm}    
\centering
    \begin{tabular}{l|r|r|r|r|r|r}
    \toprule
       Algorithm  & HalfCheetah-v2 & Humanoid-v2 & Hopper-v2 & Ant-v2  &  Walker2d-v2 & Swimmer-v2 \\
        \midrule
CoERL~(ours)  & 302.44 & 237.63 & 219.72 & 238.66 & 313.34 & 54.43\\
CoES  & 10.11 & 19.51 & 15.96 & 21.51 & 15.24 & 19.48\\
ESSAC  & 298.89 & 234.75 & 239.38 & 224.98 & 302.8 & 53.18\\
ES~(\citet{salimans2017evolution})  & 7.31 & 15.33 & 14.32 & 14.93 & 12.48 & 16.12\\
SAC~(\citet{haarnoja2018soft})  & 302.69 & 263.17 & 228.62 & 241.13 & 316.09 & 52.95\\
ERL~(\citet{khadka2018evolution})  & 1105.44 & 1309.86 & 1167.94 & 1102.4 & 1079.94 & 1028.3\\
PDERL~(\citet{bodnar2020proximal})  & 222.36 & 626.37 & 233.32 & 245.67 & 207.37 & 225.93\\
CCNCS~(\citet{yang2022evolutionary})  & 6.02 & 46.33 & 41.87 & 29.73 & 34.97 & 5.8\\
        
        \bottomrule
   \end{tabular}
    \label{tab:run_time}
\end{table*}
\subsection{Inheriting behaviour space}
\begin{figure}
    \centering
    \includegraphics[width=0.75\linewidth]{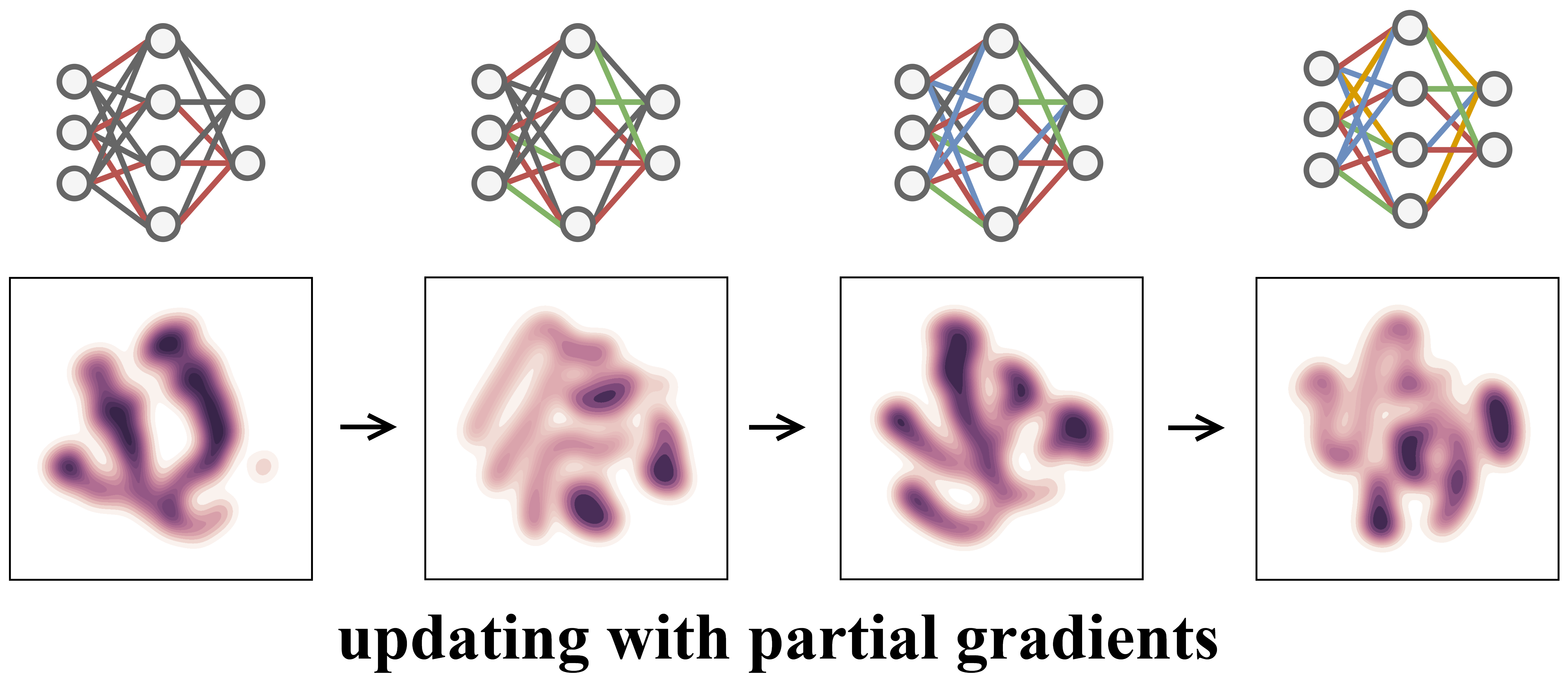}
    \caption{Four exclusive subproblems are highlighted in red, green, blue and yellow, respectively. The feature maps present the behaviour spaces, reduced by t-SNE~\cite{van2008visualizing}, of the policy after optimised in subproblems. 
    }
    \vspace{5mm}
    \label{fig:cces_behaviour}
\end{figure}

To further analyse the mechanism of CoERL, we visualise the behaviour space using t-SNE~\cite{van2008visualizing} during training. t-SNE compresses the high-dimensional observation space into the two-dimensional space. 
\emph{Swimmer-v2} is chosen for a case study.
Figure \ref{fig:cces_behaviour} shows an example of the visualisation in the case of being decomposed into four subproblems.
From left to right, the first figure is the behaviour feature map of the agent after optimising weight subset 1, then the second one shows after optimising weight subset 2, and so on. 
Each subproblem is applied for one full step before the next subproblem in CoERL, however, this update is only limited to the subproblem (denoted by parameter indices $\langle\mathcal{I}_j\rangle$).
The gradient of the preceding subproblem remains an implicit effort on the optimisation direction of the subsequent subproblem, acting as momentum. This effort is easily observed in Figure \ref{fig:cces_behaviour}, where the first feature map shares a high similarity with the third one (from left to right), as well as the second and fourth ones. This phenomenon implies that the behaviour space is inherited during optimisation.

Benefiting from the cooperative coevolution strategy and partial gradients, the offspring policy inherits the behaviour space of its parents after being updated in the last subproblem. Since CoERL only optimises the parameters of the subproblem each time, the remaining quotient set of parameters acts as a ``memory buffer". After the updates, some old neural activations of the network in the memory buffer still connect, resulting in the emergence of certain behaviours of the new policy. Additionally, the partial gradient can be considered as a form of proximal variation. Instead of aimlessly perturbing the policy, CoERL searches for the partial gradient within a proximal area, which ensures policy updates within a promising range. Then, the sampled individuals provide a certain optimisation direction, which is assembled as a vector according to Eq. \eqref{eq:e_p_g}.

\subsection{Direct coordination or indirect coordination}
A possible cause for the failure of CoERL in \emph{Swimmer-v2} could be attributed to the coordination between the CC loop and the RL loop. In ERL, the learning agent injects gradient information by replacing the worst individual in the population~\cite{khadka2018evolution}. This indirect coordination requires an extra roll-out by the learning agent itself and introduces a trade-off when choosing individuals. Our CoERL avoids these issues by sampling the population at each generation instead of maintaining a permanent population, allowing for more direct coordination. The optimised policy can directly access the evolution loop. However, we have to admit that this direct coordination may stuck in a local optimum if the base learning algorithm is not a good partner, even though CoERL has demonstrated outstanding performances on almost all tasks. Balancing the trade-off between direct and indirect coordination is worth considering as a future research direction.

\subsection{Runtime analysis}
Table ~\ref{tab:run_time} presents the average running times of algorithms in five trials. The results show that the running time of CoERL is close to SAC and much shorter than ERL.
The two loops in CoERL don't share computational resources but the evolution loop produces experiences for the RL loop.
CoERL is particularly designed for addressing the scalability issue through random grouping, which actually does not introduce additional computational complexity via random grouping. 
The complexity of the proposed partial updating is $\mathcal{O}(|\theta|)$, which should not be greater than the one-time policy gradient via backpropagation.  
CCNCS has the shortest running time as it doesn't apply gradient techniques and only updates weights a few times in each generation. However, its performance is relatively poor.

\section{Conclusion}
\label{sec:con}
In this paper, we propose CoERL, a novel cooperative coevolutionary reinforcement learning algorithm to address the scalability issues and enhance the efficiency during training. CoERL decomposes the policy optimisation problem into multiple subproblems using a cooperative coevolutionary strategy. For each subproblem, CoERL searches for partial gradients to update the policy. 
This decomposition, coupled with the use of partial gradients, ensures consistency between the behaviour spaces of parents and offspring at a reduced cost.
In contrast to traditional evolution-based approaches that discard experiences, CoERL capitalises on the collected experiences within the population, as a novel hierarchy based on cooperative coevolution. Extensive experiments on six locomotion tasks demonstrate that CoERL outperforms seven state-of-the-art algorithms and baselines. The contributions of CoERL's components are also verified through an ablation study. In the future, CoERL can be extended using knowledge-based grouping techniques and combined with other ERL algorithms like~\cite{zhu2024two} to cope with the exploration-exploitation dilemma. Furthermore, exploring the explainable decomposition in neural networks via visualisation and quantification could be another interesting direction. It is also worth investigating a thorough theoretical analysis and hyperparameter sensitivity analysis.




\bibliography{main}
\end{document}

%% file: codes/ccerl.tex
\def\old{\begin{algorithm}[h]
\caption{CoERL.}
\label{alg:ccerl}
\begin{algorithmic}[1] 
\Require The number of generations $G$, population size $\mu$, number of subproblems $m$, learning rate $\alpha$, temperature coefficient $\alpha_{s}$, noise strength $\sigma$
\Ensure Policy $\pi_\theta$
\State Initialise policy $\pi_{\theta}$, two critics $\hat{Q}_{\phi_1}$ and $\hat{Q}_{\phi_2}$
\State Initialise reward buffer $\mathcal{B}_\sR$
\For{$n=1$ to $G$}
\State Decompose the policy optimisation problem into $m$ sub-problems by grouping $|\theta|$-dimensional parameters to $m$ sets of parameter indices $\mathcal{I}_1,\cdots, \mathcal{I}_m$ 
\For{$j=1$ to $m$}
        \State Sample $\epsilon_1,\cdots,\epsilon_\mu \sim \mathcal{N}^{|\mathcal{I}_j|} (0,I)$
        \State Form the population $\{\pi_{\theta^{\langle\mathcal{I}_j\rangle}_1},\cdots,\pi_{\theta^{\langle\mathcal{I}_j\rangle}_\mu}\}$  of subproblem $\mathcal{I}_j$ by $\theta^{\langle\mathcal{I}_j\rangle}_i \leftarrow \theta^{\langle\mathcal{I}_j\rangle}+\epsilon_i$, for $i=1,\cdots,\mu$
    \For{$i=1$ to $\mu$ }
        \State $J^{\pi_{\theta_i^{\langle\mathcal{I}_j\rangle}}}_\sR,\tau^{\pi_{\theta_i^{\langle\mathcal{I}_j\rangle}}}= \text{Evaluate}(\pi_{\theta_i^{\langle\mathcal{I}_j\rangle}})$
        \State Store $\tau^{\pi_{\theta_i^{\langle\mathcal{I}_j\rangle}}}$ in $\mathcal{B}_\sR$, $f_i = J^{\pi_{\theta_i^{\langle\mathcal{I}_j\rangle}}}_\sR$
    \EndFor
        \State Update parameters of policy $\pi_\theta$ indexed by $|\mathcal{I}_j|$ : 
        \State $\theta^{\langle\mathcal{I}_j\rangle} \leftarrow \theta^{\langle\mathcal{I}_j\rangle}+\alpha\frac{1}{\mu\sigma}\sum^\mu_i f_i\epsilon_i$
\EndFor
\State Policy gradient with $\mathcal{B}_\sR$ : 
    \State Randomly sample a minibatch $B_R$ of transitions $\mathcal{T}= \langle s,a,s',r\rangle$ from $\mathcal{B}_R$
    \State Compute target $y =r-\gamma (\min\limits_{j=1,2}\hat{Q}_{\phi_j}(s',\tilde{a}')-\alpha_s \log{\pi_\theta(\tilde{a}'|s')})$, where $\tilde{a}'\sim \pi_{\theta}(\cdot|s') $
    \State Update critics with 
    \NoNumber{$\nabla_{\phi_j}\frac{1}{|B_R|}\sum\limits_{\mathcal{T} \in B_R}^{}(y-\hat{Q}_{\phi_j}(s,a))^2$ for $j=1,2$}
    \State Update actor with 
    \NoNumber{$\nabla_{\theta}\frac{1}{|B_R|}\sum\limits_{\mathcal{T} \in B_R}(\min\limits_{j=1,2}\hat{Q}_{\phi_j}(s,\tilde{a}_{\theta})-\alpha_s \log\pi_\theta(\tilde{a}_{\theta}|s))$, where $\tilde{a}_{\theta}$ is sampled from $\pi_\theta(\cdot|s)$ via the reparameterisation trick}
\EndFor
\end{algorithmic}
\end{algorithm}
}

\begin{algorithm}[h]
\caption{CoERL.}
\label{alg:ccerl}
\begin{algorithmic}[1] 
\Require The number of generations $T$, population size $\mu$, noise strength $\sigma$, number of subproblems $m$, learning rate $\alpha$, temperature coefficient $\alpha_{s}$
\Ensure Policy $\pi_\theta$
\State Initialise policy $\pi_{\theta}$, two critics $\hat{Q}_{\phi_1}$ and $\hat{Q}_{\phi_2}$
\State Initialise reward buffer $\mathcal{B}_\sR$
\For{$n=1$ to $T$}
\State Decompose the policy optimisation problem into $m$
\NoNumber{}sub-problems by grouping $|\theta|$-d parameters to $m$ sets
\NoNumber{}of parameter indices $\mathcal{I}_1,\cdots, \mathcal{I}_m$ 
\For{$j=1$ to $m$}
        \State Sample $\epsilon_1,\cdots,\epsilon_\mu \sim \mathcal{N}^{|\mathcal{I}_j|} (0,I)$
        \State Reproduce a population $\{\psi_1,\cdots,\psi_\mu\}$ for sub-
        \NoNumber{}problem $\mathcal{I}_j$ by $\psi_i \leftarrow \theta^{\langle\mathcal{I}_j\rangle}+\epsilon_i$, for $i=1,\cdots,\mu$\NoNumber{}\Comment{\emph{$\theta^{\langle\mathcal{I}_j\rangle}$ denotes $\theta$'s elements at indices $\mathcal{I}_j$}}
    \For{$i=1$ to $\mu$}
        \State $J^{\pi_{\psi_i}}_\sR,\tau^{\pi_{\psi_i}}= \text{Evaluate}(\pi_{\psi_i})$
        \State Store $\tau^{\pi_{\psi_i}}$ in $\mathcal{B}_\sR$
        \State Assign fitness $f_i = J^{\pi_{\psi_i}}_\sR$
    \EndFor
        \State $\theta^{\langle\mathcal{I}_j\rangle} \leftarrow \theta^{\langle\mathcal{I}_j\rangle}+\alpha\frac{1}{\mu\sigma}\sum^\mu_{i=1} f_i\epsilon_i$ \Comment{\emph{Eq. \eqref{eq:e_p_g}}}
        \NoNumber{}\Comment{\emph{Update parameters of policy $\pi_\theta$ indexed by $\mathcal{I}_j$}}
\EndFor
    \LineComment{\emph{Policy gradient with $\mathcal{B}_\sR$:}}
    \State Randomly sample a minibatch $\mathcal{B}$ of transitions 
    \NoNumber{}$\mathcal{T}= \langle s,a,s',r\rangle$ from $\mathcal{B}_R$
    \State Compute target 
    \NoNumber{}$y =r-\gamma (\min\limits_{j=1,2}\hat{Q}_{\phi_j}(s',\tilde{a}')-\alpha_s \log{\pi_\theta(\tilde{a}'|s')})$,
    \NoNumber{}where $\tilde{a}'\sim \pi_{\theta}(\cdot|s')$
    \State Update critics with 
    \NoNumber{}$\nabla_{\phi_j}\frac{1}{|B_R|}\sum\limits_{\mathcal{T} \in \mathcal{B}}^{}(y-\hat{Q}_{\phi_j}(s,a))^2$ for $j=1,2$
    \State Update actor with 
    \NoNumber{}$\nabla_{\theta}\frac{1}{|B_R|}\sum\limits_{\mathcal{T} \in \mathcal{B}}(\min\limits_{j=1,2}\hat{Q}_{\phi_j}(s,\tilde{a}_{\theta})-\alpha_s \log\pi_\theta(\tilde{a}_{\theta}|s))$,
    \NoNumber{}where $\tilde{a}_{\theta}$ is sampled from $\pi_\theta(\cdot|s)$ via the reparame-
    \NoNumber{}terisation trick
\EndFor

\end{algorithmic}
\end{algorithm}

%% file: tables/mujoco_5seeds_formatted.tex
\begin{table*}[t]
        \caption{Final reward performances with five different seeds in six locomotion environments. The bold number denotes the highest average value of each column. ``Avg. Rank'' denotes the average rank across all six tasks. Lower rank indicates better overall performance.
    CoERL shows the best performance with the highest average rank 1.67.}
    \vspace{3mm}
    \centering
    \begin{tabular}{l|c|c|c|c}
    \toprule
       \multicolumn{1}{c|}{\multirow{2}{*}{Algorithm}}  & HalfCheetah-v2 & Humanoid-v2 & Hopper-v2  & \multirow{2}{*}{Avg. Rank}\\
       & ~Avg. $\pm$ Std & Avg. $\pm$ Std & Avg. $\pm$ Std &  \\
        \midrule
CoERL (ours) & \textbf{11959.63 $\pm$ 250.15} & 4642.05 $\pm$ 762.38 & \textbf{3414.45 $\pm$ 100.32} & \textbf{1.67} \\
CoES & ~~~~942.63 $\pm$ 419.89 & 482.47 $\pm$ 78.36 & 991.12 $\pm$ 25.77 & 6.33 \\
ESSAC & 11597.69 $\pm$ 209.71 & ~~3931.32 $\pm$ 1309.69 & 3176.29 $\pm$ 481.59 & 3.50 \\
ES~(\citet{salimans2017evolution}) & ~~~~100.32 $\pm$ 135.40 & 490.86 $\pm$ 34.59 & 221.77 $\pm$ 41.11 & 7.67 \\
SAC~(\citet{haarnoja2018soft}) & 11774.61 $\pm$ 255.86 & \textbf{4887.86 $\pm$ 296.71} & 2842.68 $\pm$ 528.56 & 3.17 \\
ERL~(\citet{khadka2018evolution}) & ~~6790.15 $\pm$ 582.91 & ~~4677.19 $\pm$ 1054.07 & 2998.35 $\pm$ 384.52 & 3.33 \\
PDERL~(\citet{bodnar2020proximal}) & ~~7845.97 $\pm$ 285.62 & ~~765.55 $\pm$ 195.15 & 1886.00 $\pm$ 988.68 & 3.83 \\
CCNCS~(\citet{yang2022evolutionary}) & ~~~-27.73 $\pm$ 28.11 & ~~813.44 $\pm$ 161.77 & ~~384.13 $\pm$ 182.82 & 6.50 \\
    \midrule
    \midrule
       \multicolumn{1}{c|}{\multirow{2}{*}{Algorithm}}  & Ant-v2  &  Walker2d-v2 & Swimmer-v2 & \multirow{2}{*}{Avg. Rank}\\
       & Avg. $\pm$ Std & Avg. $\pm$ Std & Avg. $\pm$ Std &  \\
        \midrule
CoERL (ours) & ~\textbf{5037.22 $\pm$ 192.01} & \textbf{4962.80 $\pm$ 412.39} & 128.90 $\pm$ 40.13 & \textbf{1.67} \\
CoES & -26.53 $\pm$ 7.40 & ~~872.44 $\pm$ 116.53 & ~~55.89 $\pm$ 11.42 & 6.33 \\
ESSAC & ~~~2927.98 $\pm$ 1496.94 & ~~4228.00 $\pm$ 1283.25 & ~~94.87 $\pm$ 25.12 & 3.50 \\
ES~(\citet{salimans2017evolution}) & ~~-71.04 $\pm$ 13.72 & ~~370.08 $\pm$ 134.90 & 15.91 $\pm$ 7.77 & 7.67 \\
SAC~(\citet{haarnoja2018soft}) & ~~~3654.16 $\pm$ 1767.25 & 4397.42 $\pm$ 506.47 & ~~37.95 $\pm$ 15.59 & 3.17 \\
ERL~(\citet{khadka2018evolution}) & ~2982.24 $\pm$ 438.70 & 2790.96 $\pm$ 955.50 & 214.50 $\pm$ 85.49 & 3.33 \\
PDERL~(\citet{bodnar2020proximal}) & ~~~3730.17 $\pm$ 1484.37 & 1183.03 $\pm$ 247.26 & \textbf{305.35 $\pm$ 58.09} & 3.83 \\
CCNCS~(\citet{yang2022evolutionary}) & ~~~62.29 $\pm$ 10.66 & ~~669.85 $\pm$ 204.35 & 47.71 $\pm$ 3.01 & 6.50 \\

         \bottomrule
    \end{tabular}

    \label{tab:mujoco_5seeds}
\end{table*}